# Robust Probabilistic Inference in Distributed Systems

**Mark A. Paskin**
Computer Science Division
University of California, Berkeley

**Carlos E. Guestrin**
Berkeley Research Center
Intel Corporation

## Abstract

Probabilistic inference problems arise naturally in distributed systems such as sensor networks and teams of mobile robots. Inference algorithms that use message passing are a natural fit for distributed systems, but they must be robust to the failure situations that arise in real-world settings, such as unreliable communication and node failures. Unfortunately, the popular sum–product algorithm can yield very poor estimates in these settings because the nodes' beliefs before convergence can be arbitrarily different from the correct posteriors. In this paper, we present a new message passing algorithm for probabilistic inference which provides several crucial guarantees that the standard sum–product algorithm does not. Not only does it converge to the correct posteriors, but it is also guaranteed to yield a principled approximation at any point before convergence. In addition, the computational complexity of the message passing updates depends only upon the model, and is independent of the network topology of the distributed system. We demonstrate the approach with detailed experimental results on a distributed sensor calibration task using data from an actual sensor network deployment.

## 1 Introduction

Large-scale networks of sensing devices are a useful technology for a wide range of applications; examples include sensor networks, mobile robot teams, and certain distributed Internet applications. In these systems, nodes make local observations of their common environment in order to solve a complex global task. For example, robots in a team may each collect a set of laser scans which are combined to build a map. Or, the nodes of a sensor network in a precision agriculture deployment may collect local temperature and humidity measurements to determine when to water the crop. These local sensor measurements are often correlated, noisy or uncalibrated; as a result, probabilistic inference is of central importance to these systems.

A simple approach to inference is to download the measurements from the network and then analyze them at a central location. This approach is appropriate in some cases, but there are several reasons to prefer a distributed approach to inference. For large networks, distributed inference scales better: it reduces communication because nodes transmit compact summaries instead of their measurements; and, it leverages parallelism by making use of the computational resources at each node. In addition, distributed inference enables distributed decision-making and actuation, since every network node can access the posterior distribution of the state of the environment.

To design a distributed inference algorithm for large-scale systems there are significant challenges to overcome. First, communication between nodes can be unreliable due to noise and packet collisions, especially in ad hoc wireless networks such as those used by mobile robots and sensor networks. Second, the network topology of a distributed system can change over time; for example, communication between nearby nodes can be interrupted by occlusions or interference. Third, nodes can fail for a number of reasons, e.g., a battery may die, a computer may crash, etc. Finally, the power, computation, and communication resources of nodes can be quite limited. Because of these challenges, we have found that to solve the distributed inference problem, it is insufficient to adapt existing algorithms to distributed systems; fundamentally new algorithms are required.

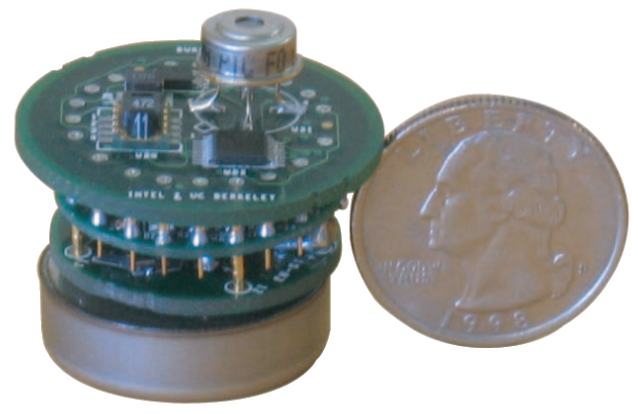

Figure 1: a MICA-2 "mote"

Sensor networks typify many of the challenges that must be overcome by a robust, distributed inference algorithm. A sensor network is a collection of autonomous devices that measure characteristics of their environment, perform local computations, and communicate with each other over a wireless network. Figure 1 shows an example sensor network node developed jointly by Intel and the University of California, Berkeley [2]; it can measure temperature, humidity, pressure, visible and infrared light, sound, magnetic fields, and acceleration. Sensor networks are an important new technology for many applications including habitat monitoring [3] and target tracking [4]. In this paper we will use sensor networks as our primary example.

A wide range of inference problems, including probabilistic inference, pattern classification, constraint satisfaction, and regression [5], can be solved by asynchronous message passing on a data structure called a junction tree [6]. In a companion paper [1], we present an architecture for distributed inference in which the nodes of the network assemble themselves into a **network junction tree**, where each network node has an associated clique and set of factors. Our architecture builds, maintains, and optimizes this network junction tree robustly, addressing both unreliable communication and node failures. Using asynchronous message passing on this junction tree, the nodes can solve the inference problem efficiently and exactly. An overview of the architecture is presented in §3.

In this architecture, each node can have an associated set of **query variables**. After the message passing algorithm converges, each node can compute the exact posterior for its query variables. However, this guarantee is of limited value: in large, lossy, or volatile networks, convergence may take a long time; or, it may never happen because the network junction tree is constantly in flux. In addition, even when a node *has* received all of its messages, *there is no way to know it;* the node can never rule out the possibility that a new node carrying additional factors will enter the network later, causing the messages to change. These problems motivate us to consider the **partial belief** a node obtains by combining its local information with its incoming messages before convergence.

Unfortunately, in the sum–product algorithm there is little we can say about the relation a node's partial belief will have to the correct posterior. For example, a missing message may carry a crucial prior factor which, when omitted, gives the partial belief a skewed view of the probability of different events. Consider a sensor network that is monitoring a nuclear reactor: if a prior factor over the boolean variable meltdown-imminent (indicating a situation that is very unlikely, *a priori*) is not integrated into a node's belief, the net effect is as if it were replaced with a uniform prior factor indicating the *a priori* chance a meltdown is imminent is 50%! As we demonstrate experimentally, this behavior makes sum–product message passing inappropriate for distributed inference in sensor networks.

Ideally, we would like an efficient, distributed message passing algorithm with three properties:

**Property 1 (Local Correctness).** *Before any communication has occurred, each node can compute the correct* **local posterior** *of its query variables given its measurements.*

**Property 2 (Global Correctness).** *After convergence, every node can compute the correct* **global posterior** *of its query variables given the measurements of all the nodes.*

**Property 3 (Partial Correctness).** *Before convergence, a node can compute the correct* **partial posterior** *of its query variables given the measurements that have been incorporated in the messages it has received.*

If every node has access to the complete probability model, then it is possible to design a simple message passing algorithm with all three properties (see §4.1). However, this approach does not scale because every node must reason with the entire model.

In this paper, we present an efficient message passing algorithm called **robust message passing**, which satisfies Properties 1 and 2, and satisfies a relaxed form of Property 3 where a node's partial posterior can make conditional independence assumptions. Thus, the algorithm is guaranteed to yield a principled approximation at any point in the inference process. In addition, robust message passing is extremely efficient: the computational complexity of the message passing updates depends only on the model, and not on the network topology. We conclude with detailed experimental results that demonstrate the algorithm on a distributed sensor calibration task using data from an actual sensor network deployment.

## 2 The distributed inference problem

We assume a network model where each node can perform local computations and communicate with other nodes over a broadcast channel. The nodes of the network may change over time: existing nodes can fail, and new nodes may be introduced. We assume a message–level error model: messages are either received without error, or they are not received at all. Only the recipient is aware of a successful transmission; neither the sender nor the recipient is aware of a failed transmission. For each pair of nodes $i$ and $j$, there is some probability that a message transmission by node $i$ will be received by node $j$; the link quality (i.e., the probability of a successful transmission) from $i$ to $j$ is unknown and may change over time, and link qualities of several node pairs may be correlated.

The random variables of our inference problem are divided into two types: observed and latent. We call the observed variables $\mathbf{M} = \{M_1, \ldots, M_K\}$ **measurement variables**; each measurement variable $M_k$ corresponds to one of the sensors on one of the nodes. We call the latent variables $\mathbf{X} = \{X_1, \ldots, X_L\}$ **environment variables**; these random variables characterize the state of the sensor network's environment. The joint probability model has two main parts. The first is a factorized prior $\Pr\{\mathbf{X}\}$ over the environment variables (such as a Bayesian network or Markov network [6]). The second part is a set of measurement models; for each measurement variable $M_k$, we have a measurement model which specifies its conditional distribution $\Pr\{M_k \mid \mathbf{B}_k\}$ given a subset of the environment variables

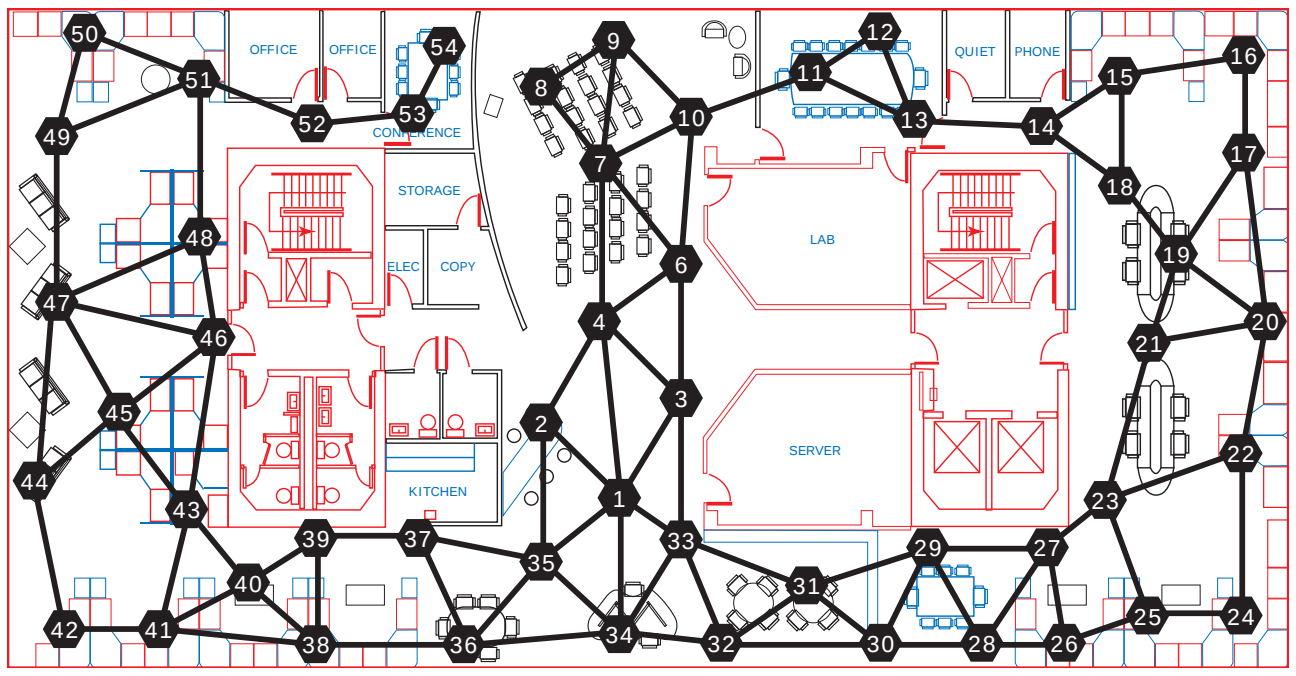

Figure 2: The Intel Berkeley Lab deployment. The Markov graph for the nodes' temperature variables is overlaid.

$\mathbf{B}_k \subseteq \mathbf{X}$. Thus, the full joint takes the form

$$\Pr\{\mathbf{X}, \mathbf{M}\} = \underbrace{\left[\frac{1}{Z}\prod_{\mathbf{C}\in\mathcal{C}} \psi_{\mathbf{C}}(\mathbf{C})\right]}_{\text{factorized prior } \Pr\{\mathbf{X}\}} \prod_{k=1}^{K} \underbrace{\Pr\{M_k \mid \mathbf{B}_k\}}_{\text{measurement model}},$$

where each $\mathbf{C} \in \mathcal{C}$ is a subset of environment variables.

We assume that each measurement model is stored by the node that obtains its corresponding measurement, and that the factors of the prior $\Pr\{\mathbf{X}\}$ are partitioned across the nodes of the network. Thus, each node $i$ receives a set of factors; collectively, these factors represent the probability model in a distributed fashion. We use $\mathbf{V}_i \subseteq \mathbf{X}$ to denote the **variables local to** $i$, i.e., the union of all environment variables associated with the factors distributed to node $i$.

In addition to its factors, each node $i$ of the network has an associated subset of environment variables called its **query variables** $\mathbf{Q}_i \subseteq \mathbf{X}$; these variables characterize the part of the environment's state that node $i$ must monitor. The **distributed inference problem** is defined as follows: after every node has obtained observations for its associated measurement variables, the nodes must collaborate so that each node $i$ obtains $\Pr\{\mathbf{Q}_i \mid \overline{m}_1, \ldots, \overline{m}_R\}$, the posterior distribution of its query variables given the measurements made by the entire network.

We now present a motivating example that will be used in the remainder of this paper.

**Example 1 (distributed sensor calibration).** *After a sensor network is deployed, the sensors can be adversely affected by the environment—for example, particulate matter can accumulate on the sensors—leading to biased measurements. The distributed sensor calibration task involves automatic detection and removal of these biases [7]. This is possible when the quantities measured by nearby nodes are correlated, but the nodes' biases are independent.*

*Figure 2 shows a sensor network of 54 nodes that we deployed in the Intel Berkeley Lab. We fit a Gaussian probability model for this data set where each node $i$ has three associated variables: its observed temperature measurement $M_i$, the true (unobserved) temperature at its location $T_i$, and the (unobserved) bias of its temperature sensor $B_i$. We assumed the temperature variables are related by the Markov graph[1] of Figure 2 and that the biases are marginally independent, resulting in a graphical model like that of Figure 3(a).*

*Given the temperature measurements, we can compute the posterior distributions of the bias variables to automatically calibrate the sensors. Under our model, we calculate that in expec-*

[1] It is important to note Figure 2 displays two things: the location of the sensor network nodes, and the graphical model for their associated temperature variables. It does *not* show the communication topology of the sensor network, which need not be related to the structure of the probability model.

*tation, the posterior temperature estimates will eliminate 44% of the bias; this estimate increases as the sensor network becomes denser, leading to more strongly correlated variables.*

*In this calibration example, the true temperature variables $T_i$ and the bias variables $B_i$ are the environment variables. (There need not be a direct correspondence between the environment variables and the sensor network nodes as in this example.) Given the graphical model, the joint probability model is*

$$\underbrace{\left[\frac{1}{Z}\prod_{(i,j)\in\mathcal{E}}\psi_{ij}(T_i,T_j)\right]}_{\textit{temperature prior}}\prod_{i\in\mathcal{N}}\underbrace{\Pr\{B_i\}}_{\textit{bias prior}}\underbrace{\Pr\{M_i \mid B_i, T_i\}}_{\textit{measurement model}},$$

*where $\mathcal{N}$ and $\mathcal{E}$ are the nodes and edges of the Markov network in Figure 2. In our calibration example, each bias prior $\Pr\{B_i\}$ is distributed to node $i$ and each binary factor of the temperature prior $\psi_{ij}(T_i, T_j)$ is distributed to node $i$ or node $j$; see Figure 3(a). (In the next section we will see that it is advantageous to distribute the factors so as to minimize the number of variables local to each node.)*

*In this task, the query variables for node $i$ are $\mathbf{Q}_i = \{T_i, B_i\}$. To solve the distributed calibration task, the nodes must collaborate so that each node $i$ obtains $\Pr\{T_i, B_i \mid \overline{m}_1, \ldots, \overline{m}_N\}$, a posterior estimate of its true temperature and bias.* □

## 3 Robust distributed inference architecture

In this section, we give an brief overview of the architecture for robust, distributed inference presented in [1]. In this architecture, the nodes of the sensor network organize themselves into a junction tree and solve the inference problem by asynchronous message passing. This is accomplished by four interacting distributed algorithms which run on each node of the network.

The first of these algorithms is **spanning tree formation**; each node in the network chooses a set of neighbor nodes so that the nodes form a spanning tree where adjacent nodes have high-quality communication links. Even when the network is fixed this is a challenging problem. The nodes of a sensor network observe only local information about the network, but spanning trees have non-local properties: they are connected; they are acyclic; and they are undirected, in that neighbors both agree that they are adjacent. In wireless sensor networks, the problem is even more difficult: link qualities are asymmetric and change over time; and, nodes must discover new neighbors and estimate their associated link qualities, as well as detect when neighbors disappear. Fortunately, spanning trees are well studied in distributed systems (e.g., for multi-hop routing in ad hoc networks), so there is a rich literature. Our spanning tree algorithm builds upon existing algorithms; it can adapt to changing network conditions, and when a stable spanning tree exists, our algorithm is guaranteed to find it.

Once a spanning tree has been constructed, the nodes have formed a distributed data structure similar to a junction tree [6]: a tree where each node $i$ has a set of variables $\mathbf{V}_i$. To make this a valid junction tree, the nodes must enforce the running intersection property: if two nodes have the same variable $X$, then all nodes on the unique path between them must also carry the variable $X$. For example, in Figure 3(c) the running intersection property does not hold because nodes 1 and 4 carry $T_2$, but node 3, which is between them in the spanning tree, does not.

Using the second algorithm of our architecture, **junction tree formation**, the nodes collaborate to learn what extra variables they must carry to enforce the running intersection property. This algorithm uses message passing along the spanning tree, much like belief propagation. For each edge $i \to j$ we define the **variables reachable to** $j$ **from** $i$ recursively by

$$\mathbf{R}_{ij} \triangleq \mathbf{V}_i \cup \bigcup_{k\in n(i)\,:\,k\neq j} \mathbf{R}_{ki},$$

where $n(i)$ are $i$'s neighbors in the spanning tree. Node $i$ computes $\mathbf{R}_{ij}$ by collecting the variables that can be reached through each neighbor but $j$ and adding its own local variables $\mathbf{V}_i$; then it sends $\mathbf{R}_{ij}$ as a message to $j$. Figure 3(d) shows three such messages for our example.

If a node receives two reachable variables messages that include some variable $X$, then it knows that it must also carry $X$. Formally, the **clique at node** $i$ is computed using

$$\mathbf{C}_i \triangleq \mathbf{V}_i \cup \bigcup_{j,k\in n(i)\,:\,j\neq k} \mathbf{R}_{ji} \cap \mathbf{R}_{ki}.$$

For example, in Figure 3(d) node 3 receives two reachable variables messages that contain $T_2$, and so it must add $T_2$ to its clique. Using these messages, node $i$ can also compute its separator with a neighbor $j$ via $\mathbf{S}_{ij} \triangleq \mathbf{C}_i \cap \mathbf{R}_{ji}$. This algorithm is guaranteed to converge to the unique set of minimal cliques that guarantee the running intersection property.

Every spanning tree induces a unique minimal junction tree for the probability model. However, some junction trees are better than others: we would like to minimize the sizes of the cliques and separators so that computation and communication are minimized. For example, if in Figure 3(d) node 4 had chosen to connect to node 1 instead of node 3, then node 3's cliques and separators would not need to include the variable $T_2$. This motivates the third algorithm of our architecture, **tree optimization**, which attempts to choose a spanning tree that induces a junction tree with small cliques and separators.

While finding the optimal spanning tree is NP-hard (by reduction from centralized junction tree optimization), we can define an efficient distributed algorithm for greedy local search through the space of spanning trees. The local move we use to move through tree space is (legal) edge swaps; in Figure 3(d) node 4 can swap its edge to 3 for an edge to 1 or 2, but node 1 cannot swap its edge to 2 for an edge to 4, because that would create a cycle. The goal is to find a spanning tree whose corresponding junction tree minimizes a cost function; this cost function can depend upon the sizes of the cliques and separators, the link qualities, and the processor power available at each node.

Nodes learn about a legal edge swap, and the change to the global cost that would occur if it was implemented, using a distributed dynamic programming algorithm. The key idea is that by starting an **evaluation broadcast** along one of its spanning tree edges, a node can learn about legal alternatives to that edge, and their relative costs. For example, in Figure 3(d), imagine that node 4 initiates an evaluation broadcast to its neighbor in the spanning tree, node 3. This request marked with the originator's identifier, as well as the identifier of its current neighbor, node 3. Node 3 then propagates the request to 1, which sees that the originator, 4, is a potential neighbor. It then sends a message to 4 outside the spanning tree, and 4 thereby learns of a legal swap: it can trade its edge to 3 for an edge to 1. By augmenting this evaluation broadcast with some compact reachable variables information, it is possible to simultaneously compute the change in global cost that would occur if the swap were executed. If the swap reduces the cost, this information is provided to the spanning tree algorithm, which effects the change.

Once a junction tree with small cliques and separators has been formed, the inference problem is solved by the fourth and final algorithm of the architecture: **belief propagation**. The nodes use the familiar sum–product message passing algorithm [6] to compute the posterior marginals of their cliques. To compute a

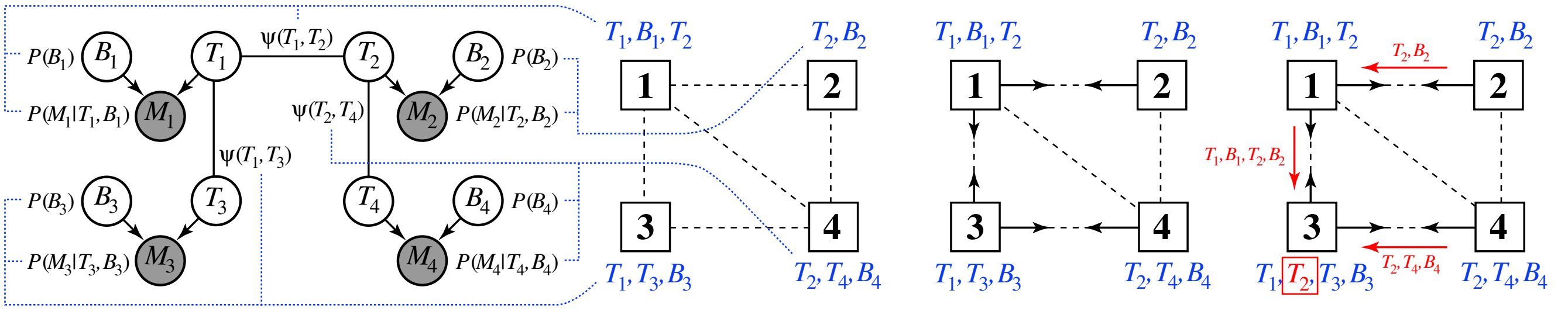

(a) distribute model factors to the sensor nodes (b) network links with good quality (c) nodes form a spanning tree (d) nodes ensure running intersection message to $j$, node $i$ computes the product of its local factors $\psi_i(\mathbf{L}_i)$ with the incoming messages from neighbors other than $j$, and marginalizes out unneeded variables:

$$\mu_{i\to j}(\mathbf{S}_{ij}) \triangleq \sum_{\mathbf{C}_i - \mathbf{S}_{ij}} \psi_i(\mathbf{L}_i) \prod_{k\in n(i)\setminus j} \mu_{k\to i}(\mathbf{S}_{ki})$$

Figure 3: Illustration of the architecture on a calibration example with four sensor nodes.

Note that node $i$ learns the separator $\mathbf{S}_{ij}$ from the junction tree formation algorithm. Once a node $i$ has received all of its messages, it can compute its **belief** as

$$\beta_i(\mathbf{C}_i) \triangleq \psi_i(\mathbf{L}_i) \prod_{k\in n(i)} \mu_{k\to i}(\mathbf{S}_{ki});$$

this belief is proportional to the posterior $\Pr\{\mathbf{C}_i \mid \mathbf{M}\}$. Our architecture uses asynchronous message passing, so that each node's belief eventually converges to the correct posterior.

In the presentation above, we made two simplifying assumptions. First, we assumed reliable communication along the edges of the spanning tree. While this is not true at the physical network layer, it can be implemented at the transport layer using message acknowledgements; by hypothesis, the spanning tree consists of high-quality wireless links. Second, we assumed that each algorithm had run to completion before the next one began; e.g., we assumed that junction tree formation begins after spanning tree formation is complete. Our algorithms cannot be implemented in this way, however, because in a sensor network, there is no way to determine when a distributed algorithm has completed: a node can never rule out the possibility that a new node will later join the network, for example.

Our algorithms therefore run concurrently on each node, responding to changes in each others' states. For example, when the spanning tree algorithm on a node adds or removes a link, the junction tree formation algorithm is informed and reacts by updating its reachable variables messages; when the junction tree formation algorithm learns that a separator has changed, it informs the belief propagation algorithm so that the messages can be updated. This tight interaction between the algorithms permits the network to react quickly when interference or node failures cause a change in the spanning tree. If the spanning tree stabilizes, then the reachable variables messages will converge, yielding a valid junction tree; eventually the belief propagation messages will also converge to the correct values, and that after this point nodes will stop passing messages.

# 4 Robust probabilistic inference

In the architecture described above, every node can compute the exact posterior for its query variables after the messages have converged. However, this guarantee is of limited value: it may take a long time for the messages to converge, or they may never converge because the network junction tree is constantly adapting to changing network conditions. Furthermore, even when a node *has* received correct versions all of its messages, *there is no way to know it;* the node can never rule out the possibility that a new node carrying additional factors will enter the network later, causing the messages (and possibly the network junction tree) to change. These problems motivate us to consider the **partial belief** a node obtains by combining its local factors with its current messages before convergence.

As we discussed in the introduction, the sum–product algorithm offers no guarantees about the relation a node's partial belief will have to the correct posterior, because before convergence a node may have failed to integrate important prior factors. This problem is especially severe in Gaussian models such as the one used in our distributed calibration example. For example, before all of a node's messages arrive, its partial belief may not be a valid density because its covariance is not positive–definite. Even if it is a valid density, missing messages can have exactly the same effect as observing a set of variables to have the value zero; in effect, the partial beliefs "hallucinate" evidence. As we demonstrate experimentally in §5, this makes partial beliefs for Gaussians densities completely unreliable.

Furthermore, the sum–product algorithm is not robust to node loss: when a node dies, it takes with it some of the factors of the probability model. The net effect is similar to what happens when there are missing messages: nodes fail to integrate factors, resulting in skewed partial beliefs. Moreover, we cannot avoid these failures by simply distributing each factor to multiple nodes, because we may multiply factors in more than once. Finally, the complexity of the sum–product algorithm scales with the sizes of the cliques of the network junction tree; thus, the computational cost depends upon the network topology and cannot be determined in advance.

In this section, we present a new message passing algorithm for probabilistic inference, called **robust message passing**, which corrects all of these problems. Not only does it converge to the correct posterior when all messages are passed, but it is also guaranteed to yield a principled approximation at any point in the inference process. In particular, it has Properties 1 and 2 (local and global correctness), and it satisfies an approximate form of Property 3 (correct partial beliefs), where the partial beliefs may make incorrect conditional independence assumptions.

In addition, the computational complexity of the message passing updates scales with the size of the model, and is independent of the network junction tree used for inference. This makes the algorithm especially attractive for use in our inference architecture; even if the network topology forces the network junction tree to have large cliques, the computational complexity of inference remains fixed. And, in contrast to the sum–product algorithm, robust message passing is correct even if each factor is distributed to several nodes, giving us a simple technique for coping with node failures.

## 4.1 A simple but impractical algorithm

We begin by first presenting a simple but impractical algorithm that has these properties, and then making it efficient. Recall that the joint probability model takes the form of a prior over the (unobserved) environment variables and a set of measurement models, one per (observed) measurement variable. In this simple algorithm, we will assume that every node of the network has access to the complete prior; as before, each node has access to its own measurement model(s). In our calibration example, every node $i$ would have the global prior $\Pr\{T_{1:N}, B_{1:N}\}$ and its measurement model $\Pr\{M_i \mid T_i, B_i\}$. In this case, Property 1 above follows easily: a node can compute its local posterior using Bayes' rule. In the calibration example, node $i$ can compute

$$\Pr\{T_{1:N}, B_{1:N} \mid \overline{m}_i\} \propto \Pr\{\overline{m}_i \mid T_i, B_i\} \times \Pr\{T_{1:N}, B_{1:N}\}$$

The local posterior $\Pr\{T_i, B_i \mid \overline{m}_i\}$ is obtained by marginalizing out all variables but $T_i$ and $B_i$.

To condition on the measurements made at other nodes, the nodes send messages to each other along the junction tree. These messages consist of likelihood functions. If node $i$ is a leaf of the junction tree, then it sends to its neighbor the likelihood of its measurements given their parent variables; in the example of Figure 3(d), node 2 sends to its neighbor the likelihood $\Pr\{\overline{m}_2 \mid T_2, B_2\}$. If node $i$ is an internal node of the junction tree, it sends to each neighbor $j$ the product of its likelihood function with the likelihood functions it receives from all neighbors but $j$. In the example of Figure 3(d), node 1 would send to node 3 the likelihood

$$\Pr\{\overline{m}_{1:2} \mid T_{1:2}, B_{1:2}\} = \underbrace{\Pr\{\overline{m}_1 \mid T_1, B_1\}}_{\text{node 1's likelihood}} \times \underbrace{\Pr\{\overline{m}_2 \mid T_2, B_2\}}_{\text{message from node 2}}$$

This product is correct because the measurements are conditionally independent given their parent variables.

To compute its belief, a node computes the product of the global prior, its local likelihood, and the likelihood messages it has received. For example, in Figure 3(d) node 3 computes its belief as

$$\Pr\{T_{1:4}, B_{1:4} \mid \overline{m}_{1:4}\} \propto \overbrace{\Pr\{T_{1:4}, B_{1:4}\}}^{\text{global prior}} \times \overbrace{\Pr\{\overline{m}_3 \mid T_3, B_3\}}^{\text{local likelihood}} \times \underbrace{\Pr\{\overline{m}_{1:2} \mid T_{1:2}, B_{1:2}\}}_{\text{message from node 1}} \times \underbrace{\Pr\{\overline{m}_4 \mid T_4, B_4\}}_{\text{message from node 4}}$$

Node 3's global posterior $\Pr\{T_3, B_3 \mid \overline{m}_{1:4}\}$ is obtained by marginalizing out all variables but $T_3$ and $B_3$. Thus, when all of the messages are passed, we obtain Property 2. Furthermore, since missing messages correspond to missing likelihoods, this algorithm also gives us Property 3; for example, if node 3 did not receive the message from node 4, its partial belief would be

$$\Pr\{T_{1:4}, B_{1:4} \mid \overline{m}_{1:3}\} \propto \overbrace{\Pr\{T_{1:4}, B_{1:4}\}}^{\text{global prior}} \times \overbrace{\Pr\{\overline{m}_3 \mid T_3, B_3\}}^{\text{local likelihood}} \times \underbrace{\Pr\{\overline{m}_{1:2} \mid T_{1:2}, B_{1:2}\}}_{\text{message from node 1}}$$

which is the posterior of the variables given the measurements incorporated in the messages, as desired.

This algorithm is impractical for two reasons. First, each node must store and reason with the complete prior, which makes the algorithm unscalable to large models. Second, the messages consist of likelihood functions over large sets of variables, which can be expensive to represent. Our solution to these problems is based upon a different representation of the global prior.

### 4.2 Decomposable reparameterization of the prior

Note that to obtain Property 1 above, each node does not need the global prior; it needs only a local prior over its query variables and the parents of its measurement variables. However, we must also maintain Property 2, which requires computation with the complete prior. One way to satisfy both of these requirements is to reparameterize the complete prior as a *decomposable probability density* [6] so it is represented in terms of a set of local priors.

For example, we can represent the prior over the temperature variables in Figure 3(a) in terms of local priors as

$$\Pr\{T_{1:4}\} = \frac{\Pr\{T_1, T_3\} \Pr\{T_1, T_2\} \Pr\{T_2, T_4\}}{\Pr\{T_1\} \Pr\{T_2\}}$$

In the general case, we must preprocess the prior to represent it as a decomposable density; this computation takes place before the factors of the model are disseminated to the nodes of the network. The first step is to form a junction tree for the original prior; we call this the **external junction tree**, to distinguish it from the network junction tree constructed in the sensor network. For example, if our probability model has the graphical model in Figure 4(a), then one possible junction tree is given in Figure 4(b). Then we use message passing on the external junction tree to compute its clique and separator marginals. The complete prior can then be represented as

$$\Pr\{\mathbf{X}\} = \frac{\prod_{\mathbf{C}} \Pr\{\mathbf{C}\}}{\prod_{\mathbf{S}} \Pr\{\mathbf{S}\}} \tag{1}$$

where $\mathbf{C}$ ranges over the cliques of the external junction tree and $\mathbf{S}$ ranges over the separators [6]. When the original prior is not decomposable, this reparameterization creates factors larger than those in the original model; for example, a decomposable representation of the temperature prior in Figure 2 has factors of up to four (rather than two) variables.

In fact, we can ignore the separator marginals in the denominator and use the clique marginals alone as an *implicit* representation of the prior. To reconstruct the full prior from this implicit representation, we would form a junction tree for the clique marginals, identify the separators, compute their marginals using the clique marginals, and then form (1). Thus, we can represent our model implicitly as a set of local priors and a set of measurement models, one per sensor.

### 4.3 Distribution of the model

Instead of giving the complete prior to each node of the network (as in the algorithm of §4.1), we will now give each node a subset of these local priors. This distributed representation of the global prior is very different from the one used by the sum–product algorithm: the prior is no longer obtained by multiplying the prior factors together; instead, it is defined implicitly by the construction of a junction tree, as described in §4.2.

This implicit representation of the global prior has several advantages. The first advantage is that we can distribute the prior factors redundantly with impugnity: if we were to reconstruct the global prior from a collection of clique marginals with replicates, the extra clique marginals would be cancelled out by extra separator marginals. This fact gives us a simple and effective technique for coping with node loss: we can distribute each factor of the model to several nodes. The factor is lost only if all of the nodes that received copies of it are lost.

Distributing the prior factors to the nodes of the network proceeds as before, with one modification. When distributing factors, we ensure that each node $i$ obtains (a copy of) the prior factors needed to compute $\Pr\{\mathbf{Q}_i, \mathbf{P}_i\}$, the joint prior over its query variables and the parents $\mathbf{P}_i$ of its measurement variables.[2] As before, the measurement models are distributed so each node obtains the measurement models for its sensors. This distribution procedure ensures Property 1:

[2]These are the prior factors that are in the smallest subtree of the external junction tree which covers $\mathbf{Q}_i \cup \mathbf{P}_i$.

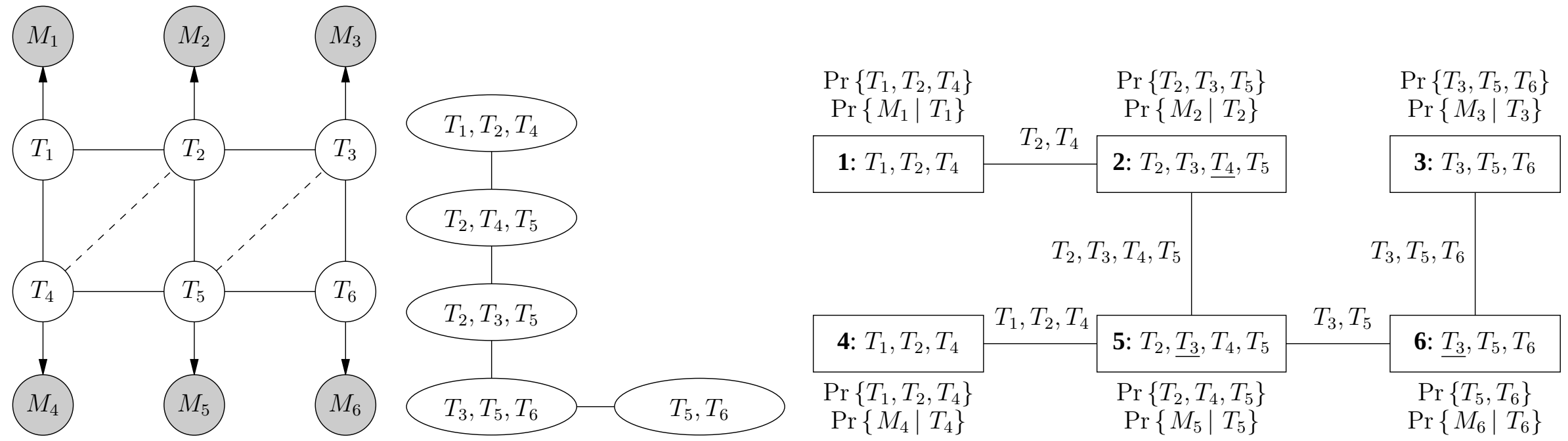

(a) graphical model (b) external junction tree (c) network junction tree and distribution of the model

Figure 4: The robust message passing example.

**Proposition 1.** *Every node can compute the local posterior of its query variables exactly without any messages.*

Figure 4(c) continues our example with a possible network junction tree; the rectangles represent the nodes of the sensor network. Each node $i$ has a copy of its measurement model $\Pr\{M_i \mid T_i\}$ as well as a prior marginal from the external junction tree. Note that each prior clique marginal has been distributed to at least one node, and that the marginal over $T_1, T_2, T_4$ has been distributed redundantly to nodes 1 and 4. In this figure, each node $i$ is labelled with its clique $\mathbf{C}_i$ (the underlined variables are added to preserve the running intersection property), and each edge is labelled with its separator.

### 4.4 Prior/likelihood factors

After the factors have been distributed to the nodes of the network, each node has a set of prior marginals and a set of measurement models. When each node obtains its observations, they are instantiated in their respective measurement models, yielding likelihood functions. The first step in our algorithm is to organize the priors and likelihoods at each node into a more convenient representation:

**Definition 1.** *A* **prior/likelihood (PL) factor** *for a set of environment variables* $\mathbf{C}$ *is a pair* $\langle \pi_{\mathbf{C}}, \lambda_{\mathbf{C}} \rangle$ *where*

- $\pi_{\mathbf{C}}$ *is a (possibly approximate) prior distribution for* $\mathbf{C}$
- $\lambda_{\mathbf{C}}$ *is a (possibly approximate) likelihood function* $\Pr\{\overline{\mathbf{m}}_{\mathbf{C}} \mid \mathbf{C}\}$ *of observations* $\mathbf{M}_{\mathbf{C}} = \overline{\mathbf{m}}_{\mathbf{C}}$ *given* $\mathbf{C}$

$\langle \pi_{\mathbf{C}}, \lambda_{\mathbf{C}} \rangle$ *is* **exact** *if* $\pi_{\mathbf{C}}$ *and* $\lambda_{\mathbf{C}}$ *are exact.*

Given the distribution of model factors described in §4.3, we can reorganize the factors allocated to node $i$ into PL factors as follows. Each measurement is instantiated in its corresponding measurement model, and the resulting likelihood is paired with the prior over the measurement's parent variables.[3] Each remaining prior factor is paired with a uniform likelihood function. Taken together, these PL factors constitute the **local model fragment** at node $i$. In the example of Figure 4(c), each node forms a single PL factor by instantiating its observation in its measurement model, and then pairing the resulting likelihood with its (only) prior factor.

The basic computations involved in robust message passing are defined in terms of combining and summarizing PL factors.

**Definition 2.** *Let* $\langle \pi_{\mathbf{C}}, \lambda_{\mathbf{C}} \rangle$ *and* $\langle \pi_{\mathbf{D}}, \lambda_{\mathbf{D}} \rangle$ *be two PL factors. The* **combination of** $\langle \pi_{\mathbf{C}}, \lambda_{\mathbf{C}} \rangle$ **and** $\langle \pi_{\mathbf{D}}, \lambda_{\mathbf{D}} \rangle$ *is*

$$\langle \pi_{\mathbf{C}}, \lambda_{\mathbf{C}} \rangle \otimes \langle \pi_{\mathbf{D}}, \lambda_{\mathbf{D}} \rangle \triangleq \left\langle \frac{\pi_{\mathbf{C}} \times \pi_{\mathbf{D}}}{\sum_{\mathbf{C}-\mathbf{D}} \pi_{\mathbf{C}}}, \lambda_{\mathbf{C}} \times \lambda_{\mathbf{D}} \right\rangle.$$

[3]To simplify the exposition, we will assume the external junction tree is chosen so that for each measurement variable, there is a clique that covers the measurement variable's parents.

Using the product rule for probabilities, it is easy to prove

**Proposition 2.** *The combination rule of Definition 2 is exact when* $\langle \pi_{\mathbf{C}}, \lambda_{\mathbf{C}} \rangle$ *and* $\langle \pi_{\mathbf{D}}, \lambda_{\mathbf{D}} \rangle$ *are exact and we have*

$$\mathbf{C} \cup \mathbf{M}_{\mathbf{C}} \perp\!\!\!\perp \mathbf{D} \cup \mathbf{M}_{\mathbf{D}} \mid \mathbf{C} \cap \mathbf{D}, \tag{2}$$

*where* $\mathbf{A} \perp\!\!\!\perp \mathbf{B} \mid \mathbf{C}$ *means* $\mathbf{A}$ *and* $\mathbf{B}$ *are conditionally independent given* $\mathbf{C}$*.*

When these conditions do not hold the combination rule is approximate (and perhaps asymmetric, since we may have $\sum_{\mathbf{C}-\mathbf{D}} \pi_{\mathbf{C}} \neq \sum_{\mathbf{D}-\mathbf{C}} \pi_{\mathbf{D}}$).

The other operation on PL factors is summarization:

**Definition 3.** *Let* $\langle \pi_{\mathbf{D}}, \lambda_{\mathbf{D}} \rangle$ *be a PL factor and* $\mathbf{S}$ *be a set of random variables. The* **summary of** $\langle \pi_{\mathbf{D}}, \lambda_{\mathbf{D}} \rangle$ **to** $\mathbf{S}$ *is*

$$\bigoplus_{\mathbf{S}} \langle \pi_{\mathbf{D}}, \lambda_{\mathbf{D}} \rangle \triangleq \left\langle \sum_{\mathbf{D}-\mathbf{S}} \pi_{\mathbf{D}}, \frac{\sum_{\mathbf{D}-\mathbf{S}} \pi_{\mathbf{D}} \times \lambda_{\mathbf{D}}}{\sum_{\mathbf{D}-\mathbf{S}} \pi_{\mathbf{D}}} \right\rangle.$$

This summary rule simply computes a marginal of the prior, and computes the marginal likelihood by forming the joint, marginalizing it down, and dividing out the marginal prior. The summary rule is exact when the inputs are exact.

### 4.5 Robust message passing

Given this reparameterization, each node has a set of PL factors which represent a fragment of the complete posterior model. If we were to assemble all of the nodes' fragments in one location, we could form the posterior joint density and solve the inference problem. Instead, we will develop a message passing algorithm which interleaves assembly of the model with inference, so that the nodes can use dynamic programming to compute the posterior marginals they need efficiently.

Just like other junction tree message passing algorithms, this algorithm uses combine and summary (collapse) operations on a factor representation. (In the sum–product algorithm, the combine operation is multiplication, the summary operation is marginalization, and the factors are potential functions.) As a result, we can use the inference architecture presented in §3 without change. In our new message passing algorithm, each factor is a collection of prior/likelihoods:

**Definition 4.** *A* **model fragment factor** $\Phi$ *is a collection of PL factors* $\{\langle \pi_{\mathbf{C}}, \lambda_{\mathbf{C}} \rangle : \mathbf{C} \in \mathcal{C}\}$*.* $\Phi$ *is* **exact** *if all of its member PL factors are exact.*

Model fragment factors represent both forms of factorization described above: the prior and likelihood information are kept separate, and the prior is represented implicitly in terms of a collection of local priors. Recall that assembling the local marginals of a decomposable density into the complete density requires building a junction tree; in a similar fashion, computing the posterior distribution represented by a model fragment factor also involves building a clique tree.

**Definition 5.** *Let $\Phi$ be a model fragment factor. A* **canonical clique tree** *for $\Phi$ is a tree over the* PL *factors of $\Phi$ which has maximum cardinality variable intersections of neighboring cliques. $\Phi$ is* **consistent** *if it has a canonical clique tree $\mathcal{T}$ such that (1) $\mathcal{T}$ satisfies the running intersection property and (2) the conditional independencies encoded by $\mathcal{T}$ are also encoded by the external junction tree.*

(Note that we explicitly allow for the possibility that a canonical clique tree does not have the running intersection property; this fact will be important later when we discuss partial beliefs.) Using a canonical clique tree, we can flatten a model fragment into a single PL factor:

**Definition 6.** *Let $\Phi = \{\langle \pi_{\mathbf{C}}, \lambda_{\mathbf{C}} \rangle : \mathbf{C} \in \mathcal{C}\}$ be a model fragment factor. A* PL *factor $\langle \pi_{\mathbf{V}}, \lambda_{\mathbf{V}} \rangle$ is a* **flattening of** $\Phi$ *if it can be obtained by the following procedure:*

1. *Compute a canonical clique tree $\mathcal{T}$ for $\Phi$.*
2. *Repeat: let $\mathbf{C}$ be a leaf clique of $\mathcal{T}$ with neighbor $\mathbf{D}$. Replace $\mathbf{C}$ and $\mathbf{D}$ with a new clique $\mathbf{C} \cup \mathbf{D}$ whose associated* PL *factor is $\langle \pi_{\mathbf{D}}, \lambda_{\mathbf{D}} \rangle \otimes \langle \pi_{\mathbf{C}}, \lambda_{\mathbf{C}} \rangle$.*

The posterior represented by $\Phi$ is then computed as $\pi_{\mathbf{V}} \times \lambda_{\mathbf{V}}$, the product of the "flat" prior and likelihood.

When $\Phi$ contains enough of the PL factors of the original model to satisfy the consistency property, and all of the PL factors in $\Phi$ are exact, then the flattening is also exact (and therefore unique):

**Proposition 3.** *Let $\Phi$ be an exact and consistent model fragment factor, then the flattening of $\Phi$ produces an exact* PL *factor.*

This is proved by verifying that the canonical clique tree guarantees the conditional independencies required for the PL combinations to be exact.

We now define the combine and summary operators for model fragment factors; these are the main operations of robust message passing. The combine operator is simply union: combining two model fragments results in a new model fragment with the union of their member PL factors. (As an optimization, we may eliminate non-maximal PL factors in the result by combining them with a subsuming PL factor using Definition 2.) The summary operation is where the work of inference is done:

**Definition 7.** *Let $\Phi = \{\langle \pi_{\mathbf{C}}, \lambda_{\mathbf{C}} \rangle : \mathbf{C} \in \mathcal{C}\}$ be a model fragment factor and let $\mathbf{S}$ be a set of random variables. Another model fragment factor $\Psi$ is a* **summary of** $\Phi$ **to** $\mathbf{S}$ *iff it can be obtained by the following procedure:*

1. *Compute a canonical clique tree $\mathcal{T}$ for $\Phi$.*
2. *Repeat: let $\mathbf{C}$ be a leaf clique of $\mathcal{T}$ with neighbor $\mathbf{D}$ such that $\mathbf{C} \cap \mathbf{S} \subseteq \mathbf{D}$. (If there is no such clique, terminate.) Update the* PL *factor of $\mathbf{D}$ as follows:*

$$\langle \pi'_{\mathbf{D}}, \lambda'_{\mathbf{D}} \rangle = \langle \pi_{\mathbf{D}}, \lambda_{\mathbf{D}} \rangle \otimes \left( \bigoplus_{\mathbf{C} \cap \mathbf{D}} \langle \pi_{\mathbf{C}}, \lambda_{\mathbf{C}} \rangle \right) \qquad (3)$$

*Then remove $\mathbf{C}$ from $\mathcal{T}$ and $\langle \pi_{\mathbf{C}}, \lambda_{\mathbf{C}} \rangle$ from $\Phi$.*

Informally, this summary operation repeatedly prunes a PL factor whose prior is no longer needed by "transferring" its likelihood information onto another PL factor. (It is also possible to prune "internal" nodes in the model fragment factor; we omit details due to lack of space.) When $\Phi$ is exact and contains enough of the PL factors to guarantee consistency with respect to every eliminated variable, then the summary is also exact:

**Proposition 4.** *Let $\Phi$ be an exact model fragment factor. Then any summary of $\Phi$ to $\mathbf{S}$ is exact if for any* PL *factor $\langle \pi_{\mathbf{S}}, \lambda_{\mathbf{S}} \rangle$, $\Phi \cup \{\langle \pi_{\mathbf{S}}, \lambda_{\mathbf{S}} \rangle\}$ is a consistent model fragment factor.*

This proposition is proved by verifying that the canonical clique tree of $\Phi \cup \{\langle \pi_{\mathbf{S}}, \lambda_{\mathbf{S}} \rangle\}$ guarantees the conditional independencies required for the PL combinations to be exact.

**Example 2 (robust message passing).** *To illustrate the robust message passing algorithm, we describe the computation of the message from node 5 to node 6 in Figure 4(c). Node 5's local model fragment has only one* PL *factor:*

$$\Phi_5 = \{\langle \Pr\{T_2, T_4, T_5\}, \Pr\{\overline{m}_5 \mid T_2, T_4, T_5\} \rangle\}$$

*Node 5 computes its message to node 6 by combining this local model fragment with the messages it receives from nodes 2 and 4. These messages are given by*

$$\begin{aligned} \Psi_{2\to 5} = \{ & \langle \Pr\{T_1, T_2, T_4\}, \Pr\{\overline{m}_1 \mid T_1, T_2, T_4\} \rangle, \\ & \langle \Pr\{T_2, T_3, T_5\}, \Pr\{\overline{m}_2 \mid T_2, T_3, T_5\} \rangle \} \\ \Psi_{4\to 5} = \{ & \langle \Pr\{T_1, T_2, T_4\}, \Pr\{\overline{m}_4 \mid T_1, T_2, T_4\} \rangle \} \end{aligned}$$

*To combine these model fragments, we simply compute the union of the* PL *factors above. The message from node 5 to node 6 is the summary of this combination down to the separator between nodes 5 and 6: $\mathbf{S}_{56} = \{T_3, T_5\}$. From Definition 7, the first step in computing a summary is forming a canonical clique tree for the* PL *factors; one such clique tree is shown in Figure 5(a). The variables in $\mathbf{S}_{56}$ are underlined.*

*The next step is to iteratively identify leaf* PL *factors whose prior information can be discarded, and to "transfer" their likelihood information onto retained* PL *factors. In Figure 5(a) there are two leaves, but only the bottom one can be pruned, as its does not overlap with $\mathbf{S}_{56} = \{T_3, T_5\}$, the variables of interest. To prune this leaf, we use Equation* (3) *to transfer the leaf's likelihood to its neighbor; in this case the likelihoods are simply multiplied together because the cliques are the same. Then the leaf is removed, yielding the clique tree of Figure 5(b), in which a new leaf is exposed. Because this new leaf has no intersection with $\mathbf{S}_{56}$, it too can be pruned. In this case, we use Equation* (3) *to update the neighbor's* PL *factor to*

$$\begin{aligned} & \left\langle \begin{matrix} \Pr\{T_2, T_4, T_5\}, \\ \Pr\{\overline{m}_5 \mid T_2, T_4, T_5\} \end{matrix} \right\rangle \otimes \bigoplus_{\{T_2, T_4\}} \left\langle \begin{matrix} \Pr\{T_1, T_2, T_4\}, \\ \Pr\{\overline{m}_1, \overline{m}_4 \mid T_1, T_2, T_4\} \end{matrix} \right\rangle \\ &= \left\langle \begin{matrix} \Pr\{T_2, T_4, T_5\}, \\ \Pr\{\overline{m}_5 \mid T_2, T_4, T_5\} \end{matrix} \right\rangle \otimes \left\langle \begin{matrix} \Pr\{T_2, T_4\}, \\ \Pr\{\overline{m}_1, \overline{m}_4 \mid T_2, T_4\} \end{matrix} \right\rangle \\ &= \left\langle \begin{matrix} \Pr\{T_2, T_4, T_5\}, \\ \Pr\{\overline{m}_1, \overline{m}_4, \overline{m}_5 \mid T_2, T_4, T_5\} \end{matrix} \right\rangle \end{aligned}$$

*and prune the leaf to yield the clique tree of Figure 5(c). Even though the bottom clique contains $T_5$ (which is in $\mathbf{S}_{56}$), this clique can still be pruned by the rule in Definition 7; intuitively, the prior information this clique represents about $\mathbf{S}_{56}$ is redundant because it is also represented by the top clique. Transferring the likelihood information to the top clique yields the final clique tree shown in Figure 5(d). This single* PL *factor is the message that is sent from node 5 to node 6.* □

When the network junction tree has the running intersection property, Proposition 4 can be used to prove that all of the summaries performed by synchronous robust message passing yield correct results. This gives us Property 2 (global correctness)

**Proposition 5.** *When asynchronous robust message passing is used on a valid network junction tree, each node's belief converges to a model fragment factor representing the correct global posterior.*

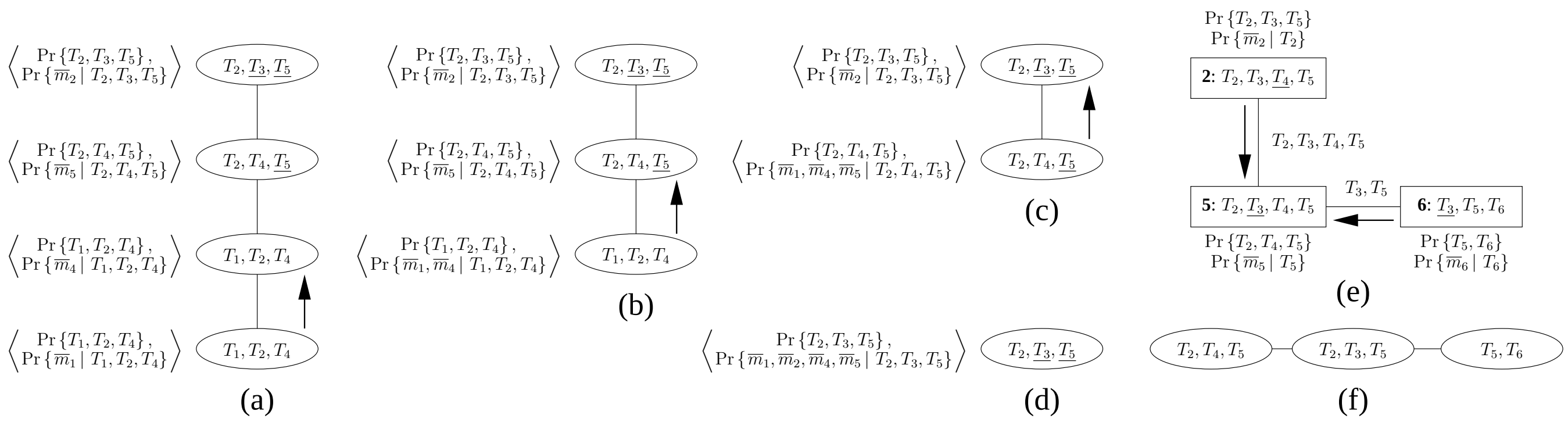

Figure 5: Robust message passing example, continued.

In our inference architecture, this result translates into the following guarantee: if the environment is such that a stable spanning tree can be built, the junction tree formation algorithm will eventually construct a valid junction tree and every node's belief will converge to the correct global posterior.

In addition to being correct, this robust message passing algorithm is efficient. Recall that all of the computations performed in robust message passing are combinations and summaries of PL factors which were derived from the external junction tree. In some sense, the nodes of the network junction tree are performing message passing on the external junction tree; as a result, the complexity of robust message passing is determined by the width of the external junction tree, not that of the network junction tree. This is an invaluable property for a distributed inference algorithm: *the computational complexity is determined by the model, and not by the communication topology.* However, this does not mean that the network junction tree plays no role in the inference algorithm. The cliques and separators of the network junction tree determine when it is safe to prune PL factors out of a model fragment; if a poor network junction tree is selected, the nodes of the network junction tree must communicate to each other more pieces of the external junction tree to solve the inference problem. Thus, the spanning tree optimization algorithm described in §3 is still needed, but only to reduce the communication cost of inference, not the computational complexity.

### 4.6 Partial beliefs

Unlike the sum–product message passing algorithm, our robust message passing algorithm makes it easy to characterize the partial belief a node forms before it has received exact versions of all of its messages. The fundamental computations involved in robust message passing are combinations and summaries of PL factors. These operation preserve exactness with one exception: when we combine two PL factors and the conditional independence (2) does not hold. Therefore, inexactness arises in the form of incorrect conditional independence assumptions.

For example, consider the situation in Figure 5(e), where node 5 has received messages only from nodes 2 and 6 (neither of which have received messages from their other neighbors). In this case, node 5 receives all factors at nodes 2 and 6, but it does not have access to a prior factor stored on node 3. When it computes its partial belief using Definition 6, the clique tree it forms (shown in Figure 5(f)) is approximate: it represents the conditional independence $T_3 \perp\!\!\!\perp T_6 \mid T_5$, which is incorrect given the original model in Figure 4(a). When the likelihood terms are incorporated, additional incorrect conditional independence assumptions are made (e.g., that $M_3 \perp\!\!\!\perp M_6 \mid T_5$).

When the network junction tree is valid (in that it has the running intersection property), approximation arises only from these conditional independence assumptions (not by summing variables out early). In this case, the messages used (directly or indirectly) to compute a node's partial belief may encode incorrect conditional independence assumptions, but they are consistent in these assumptions in the following sense:

**Proposition 6.** *Let $i$ be a node of a valid network junction tree in which some, but not all, of the robust messages have been passed. Then there exists a set of conditional independencies $\mathbb{I}$ such that the partial belief of node $i$ is exact under the model*

$$\tilde{P}(\mathbf{X}, \mathbf{M}) = \underset{\widehat{P}(\mathbf{X},\mathbf{M}) \models \mathbb{I}}{\operatorname{argmin}} \; D(P(\mathbf{X}, \mathbf{M}) \,\|\, \widehat{P}(\mathbf{X}, \mathbf{M})),$$

*which minimizes the Kullback–Liebler divergence from the true model given the conditional independences $\mathbb{I}$.*

This is a relaxed form of Property 3 because the partial posterior may make incorrect independence assumptions.

When the network junction tree is not valid, then variables may be summed out too early in the computation of the messages, which can cause the messages used in computing a node's partial belief to make different conditional independence approximations. In this case, a node's partial belief is not exact under any fixed approximate model: likelihoods have been incorporated with inconsistent conditional independence assumptions in the prior. As our experimental results demonstrate, even in this case the partial beliefs can be excellent approximations.

## 5 Experiments

To validate our algorithms we have tested them on the sensor calibration task presented in the introduction. We deployed 54 Intel–Berkeley motes in our lab (Figure 2) and collected temperature measurements every 30 seconds for a period of several days. We also collected link quality statistics and computed the fraction of transmissions each mote heard from every other mote. Using this link quality information, we designed a sensor network simulator that modeled the actual deployment. (Designing, testing, and experimenting with our algorithms would have been far more difficult in the actual deployment.) This simulator uses an event-based model to simulate lossy communication between the nodes at the message level: messages are either received or not, depending upon samples from the link quality model. The simulator's programming model is also event-based—algorithms are coded in terms of responses to message events—and we expect that our implementations can be transferred to a real sensor network without significant changes.

Using the temperature data we fit a Gaussian distribution with the Markov graph depicted in Figure 2. (Mote 5 failed shortly after deployment, which explains its absence in the figure, and also justifies our efforts to develop algorithms robust to such failures.) The model was augmented with bias variables for each temperature measurement, which were distributed i.i.d. from $\mathcal{N}(0, 1°\mathrm{C})$. We then sampled a true, unobserved bias for each node, and created a set of biased measurements by adding these biases to a held-out test set of measurements. The task is for the nodes to compute their posterior mean biases, and the error metric we use is the root mean squared error (RMS) from their estimates to the (unobserved) biases we sampled.

### 5.1 Convergence to optimal global inference

Our first experiment demonstrates the inference architecture in the simplest setting, where link qualities are stable. Figures 6(a) and 6(b) visualize traces of the inference architecture when robust and sum–product message passing are used; the spanning tree and junction tree formation algorithms are identical in both cases (for simplicity the optimization algorithm was not used). The $x$-axis of all of these plots is time. At the bottom of each trace are two bars: the bottom bar is white when a valid spanning tree has been constructed, and the top bar is white when the running intersection property has been enforced. At the beginning of these simulations there is a prolonged period before the spanning tree is built; the nodes wait until they have accurate link quality estimates before they begin building the spanning tree. Notice that after a spanning tree is built, it can be lost; this typically occurs in the delay between a node swapping a neighbor and the old and new neighbor learning of the change. Also note that after a stable spanning tree has been found, the junction tree formation algorithm eventually enforces the running intersection property, resulting in a valid junction tree.

The main panel of Figure 6(a) plots the RMS error of three inference algorithms. The line marked *global* refers to centralized inference using all of the measurements. In this case, the posterior mean bias estimates of global inference have 0.61 RMS error; because the bias is additive, this number also represents the average error in the posterior mean temperature measurements. The line marked *local* refers to centralized local inference, where each node's posterior is computed using only its measurement. Local inference performs about as well as predicting zero bias, achieving a 0.99 RMS error; this is expected, since solving the calibration problem requires correlated measurements at different nodes.

The third curve, *distributed robust*, refers to the robust message passing algorithm. This plot graphically demonstrates the key properties of the algorithm: before any messages have been passed, the partial beliefs coincide with the estimates given by centralized local inference; at convergence, the estimates coincide with those of centralized global inference; and, before all messages have been passed, the estimates are informative approximations. Looking closely, we can see that before the junction tree is valid, and even before a complete spanning tree is constructed, the estimates of the robust message passing algorithm quickly approach those of centralized global inference.

We now turn our attention to Figure 6(b), which was generated from the same setup as Figure 6(a) except that the sum–product message passing algorithm was used. Here too we see that the message passing algorithm converges to the correct global posteriors once a valid junction tree is formed. But before convergence, its estimates are extremely poor—many times worse than local inference. Moreover, its convergence is not gradual, but quite sudden: the estimates are useless right up to the moment of convergence. Finally, the top panel shows the number of nodes whose partial beliefs are not valid densities; before convergence, many nodes are unable to make a prediction at all. Taken together, these qualitative behaviors make sum–product message passing inappropriate for distributed inference in sensor networks.

### 5.2 Optimization of the spanning tree

Using the same setup as §5.1, we ran an experiment to test the distributed spanning tree optimization algorithm (Figure 6(c)). We chose our communication cost function so that the cost of a (directed) edge is proportional to the expected number of transmitted bytes necessary to successfully communicate a belief propagation message; this cost function takes into account the link quality of an edge as well as the size of its separator. Offline, we used a combination of simulated annealing, greedy local search, and random restarts to find a local minimum of this cost function[4]; its cost is plotted as the horizontal line in Figure 6(c). The piecewise constant curve represents the current cost of the spanning tree, when one exists. Note that the initial spanning tree, which is selected using only link quality information, is significantly more expensive than the hypothesized optimum, but that the distributed optimization algorithm eventually finds trees whose cost is less than twice this hypothesized optimum.

### 5.3 Robustness to communication failure

To test the algorithms' robustness to long-term communication failure, we ran the experiment of §5.1 again, but this time we introduced a period where interference causes the network to be segmented into two parts. At time 60, all messages between the left half of the network (nodes 1-35) and the right half (nodes 36-54) are lost; at time 120 the communication is restored. During the interference the nodes on the left half of the network do not have access to the measurements made on the right, and vice-versa; therefore the global inference error curve in Figure 6(d) changes: it is computed for each node using the posterior conditioned on the measurements on its side of the network.

In Figure 6(d) we can see that the robust message passing algorithm achieves convergence before and after the inferference period, but that the interference prevents a (complete) spanning tree from being formed. In spite of this, the robust message passing algorithm converges with an error that is very close to the optimum. In this period, each half of the network forms its own junction tree and uses message passing. Robust message passing does not converge to exactly the same result as global inference because some prior factors needed by the left half of the network have been distributed to right half, and vice versa.

We also ran the same experiment using sum–product message passing. Like robust message passing, it converged to the correct belief before and after the interference period; however, during the interference both halves of the network converged to very poor estimates (RMS error 1.95). This is expected because during the period of interference, each side of the network is missing half of the prior model.

### 5.4 Robustness to node failures

In this experiment we used the setup from §5.1 with simulated node failures. The time to failure for each node was sampled i.i.d. from an exponential distribution. As each node dies, its measurement is lost, and so the inference problem to be solved is changing over time; this explains the changing error values for global and local inference.

We ran this experiment with both the sum–product and robust message passing algorithms (with the same sampled failure schedule). The sum–product algorithm failed miserably because the first node died at time 6.86, before a junction tree could even be formed. As a result, prior factors were lost before message passing could even converge. The RMS error was never better than that of local inference, and for the majority of the simulation, it was far worse.

Figures 6(e) and 6(f) show how robust message passing performs in the presence of failing nodes. In Figure 6(e), each prior factor was distributed to only one node of the network; in this case we see that robust message passing gives consistently reasonable results; it converges to the global optimum until approximately time 200, when it no longer has all of the factors to reconstruct the true prior. After this point its performance gracefully degrades to the correct results of local inference. Figure 6(f) shows that when each factor is distributed redundantly to three nodes, the algorithm is still correct, and for much longer: it solves the global inference problem exactly past 500 seconds, when only 26 of the original 53 nodes are still functioning.

[4]The local moves used in this optimization were from a larger class of edge swaps than that used by the distributed algorithm.

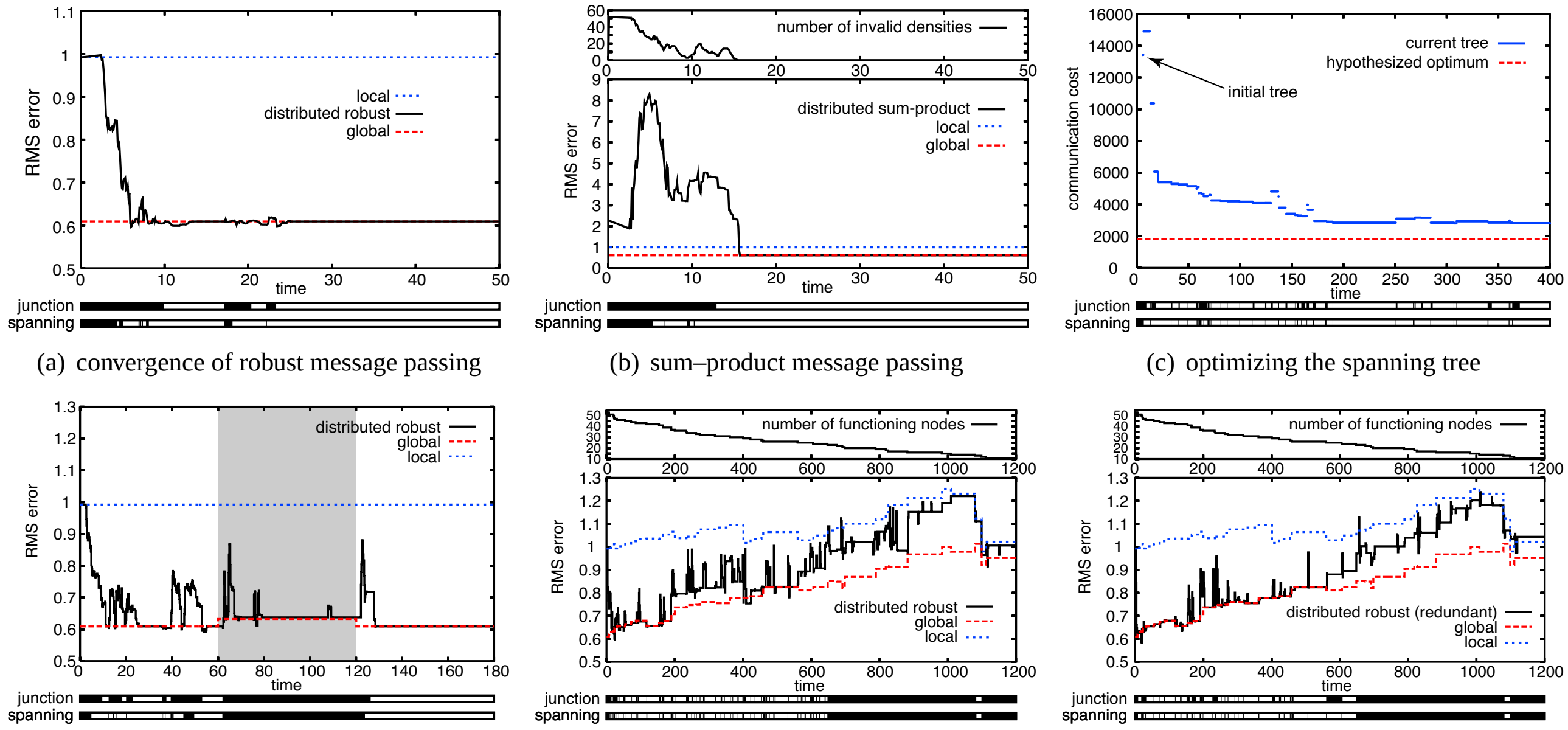

(a) convergence of robust message passing (b) sum–product message passing (c) optimizing the spanning tree

(d) coping with interference (e) failing nodes (f) redundant distribution of factors

Figure 6: Experimental results.

### 5.5 Applying loopy belief propagation

Because it is simple to implement and easily parallelizable, it has been suggested that loopy belief propagation (LBP) is "an ideal computational and communication framework for sensor networks" [8]. We had hoped to compare our algorithms to LBP, but unfortunately, this was not possible; LBP can only be applied to a restricted class of models, and the model we learned from the sensor network data was not in this class.[5] Even so, we can infer some qualitative properties of LBP from our experiments with its exact correlate, the sum–product algorithm. Like the sum–product algorithm, LBP can perform poorly when factors of the prior model are inaccessible due to node failure or interference. For the same reason, we could not expect LBP to provide reasonable answers until enough time had elapsed to permit messages from one end of the network to reach the other.

## 6 Conclusions

Distributed systems are an exciting application area for probabilistic reasoning algorithms with new and complex challenges to overcome. To address these challenges, we found that it is insufficient to adapt existing algorithms; for example, we have demonstrated that sum–product message passing can fail badly when communication is unreliable or nodes fail. In this paper, we presented a novel, robust algorithm for probabilistic inference in distributed systems that provides very strong theoretical guarantees: even when nodes fail and messages are lost, each node can compute a principled approximation of the posterior distribution of its query variables, given the measurements incorporated in the messages the node has received. In addition, robust message passing is extremely efficient: the computational complexity of the message passing updates depends only on the model, and not on the underlying network topology. We also demonstrated the theoretical properties of our algorithm with detailed experimental results on data from an actual sensor network deployment.

This paper focuses on static inference problems, but many distributed systems problems require reasoning about time with dynamic Bayesian networks (DBN). Unfortunately, exact DBN inference can be intractable. In current work, we are designing a new inference algorithm that addresses both the robustness requirements of distributed systems, and the computational complexity of inference in dynamic probability models.

[5] LBP can only be applied to Gaussian models where all of the pairwise factors are normalizable; otherwise the messages and beliefs may not be well-defined. The model we learned from data cannot be expressed in this way; even a modified version of LBP failed to converge.

### Acknowledgements

We gratefully acknowledge Wei Hong, Samuel Madden, and Romain Thibaux for their help deploying the sensor network. Mark Paskin was supported by ONR N00014-00-1-0637 and an Intel Research Internship.